\documentclass{article}
\usepackage{ijcai26}
\usepackage{tabularx}
\usepackage{times}
\usepackage{soul}
\usepackage{url}
\usepackage[hidelinks]{hyperref}
\usepackage[utf8]{inputenc}
\usepackage[small]{caption}
\usepackage{graphicx}
\usepackage{amsmath}
\usepackage{amsthm}
\usepackage{booktabs}
\usepackage{algorithm}
\usepackage[noend]{algpseudocode}
\usepackage[switch]{lineno}
\usepackage{booktabs}   
\usepackage{multirow}   
\usepackage{graphicx}
\usepackage{caption}
\usepackage{subcaption}
\usepackage{tabularx}  
\usepackage[strings]{underscore}
\nolinenumbers

\title{UP-NRPA: User Portrait based Nested Rollout Policy Adaptation for Planning with Large Language Models in Goal-oriented Dialogue Systems}

\author{
Hui Wang$^{1,2}$\and
Fafa Zhang$^{1,2}$\and
Meng Liu$^{1,2}$\and
Xiangyu Chen$^{1,2}$\and
Chaoxu Mu$^{1,2,3}$\\
\affiliations
$^1$School of Artificial Intelligence, Anhui University\\
$^2$Anhui Provincial Key Laboratory of Security Artificial Intelligence, Anhui University\\
$^3$Pengcheng Laboratory, Shenzhen, China\\
\emails
 h.wang.13@ahu.edu.cn, \{wa23301160, w125221177, xiangyu0113\}@stu.ahu.edu.cn, cxmu@tju.edu.cn
}
\begin{document}

\maketitle

\begin{abstract}
To address the challenge that current dialogue policy planning methods struggle to dynamically adapt to diverse user characteristics, this paper proposes a User Portrait based Nested Rollout Policy Adaptation (UP-NRPA) online framework with Large Language Models. In contrast to conventional approaches dependent on model training and require offline reinforcement learning policy models for user groups, UP-NRPA enables dynamic customization of dialogue strategies through an adaptive mechanism. This is achieved by leveraging real-time user feedback alongside personality, preferences, and objectives mapped from the current user portrait, thereby adapting to user characteristics without offline reinforcement learning. In collaborative and non-collaborative dialogue benchmarks, UP-NRPA demonstrated considerable benefits, achieving an impressive 100\% success rate in multiple dialogue tasks. Particularly in negotiation tasks, the sale-to-list ratio (SL) increased by 56.41\%. This demonstrates that UP-NRPA can adapt to diverse user needs without requiring a training mechanism, enabling the dialogue system to adapt to user characteristics.
\end{abstract}

\begin{figure*}[htbp]
    \centering
    \includegraphics[width=0.995\textwidth]{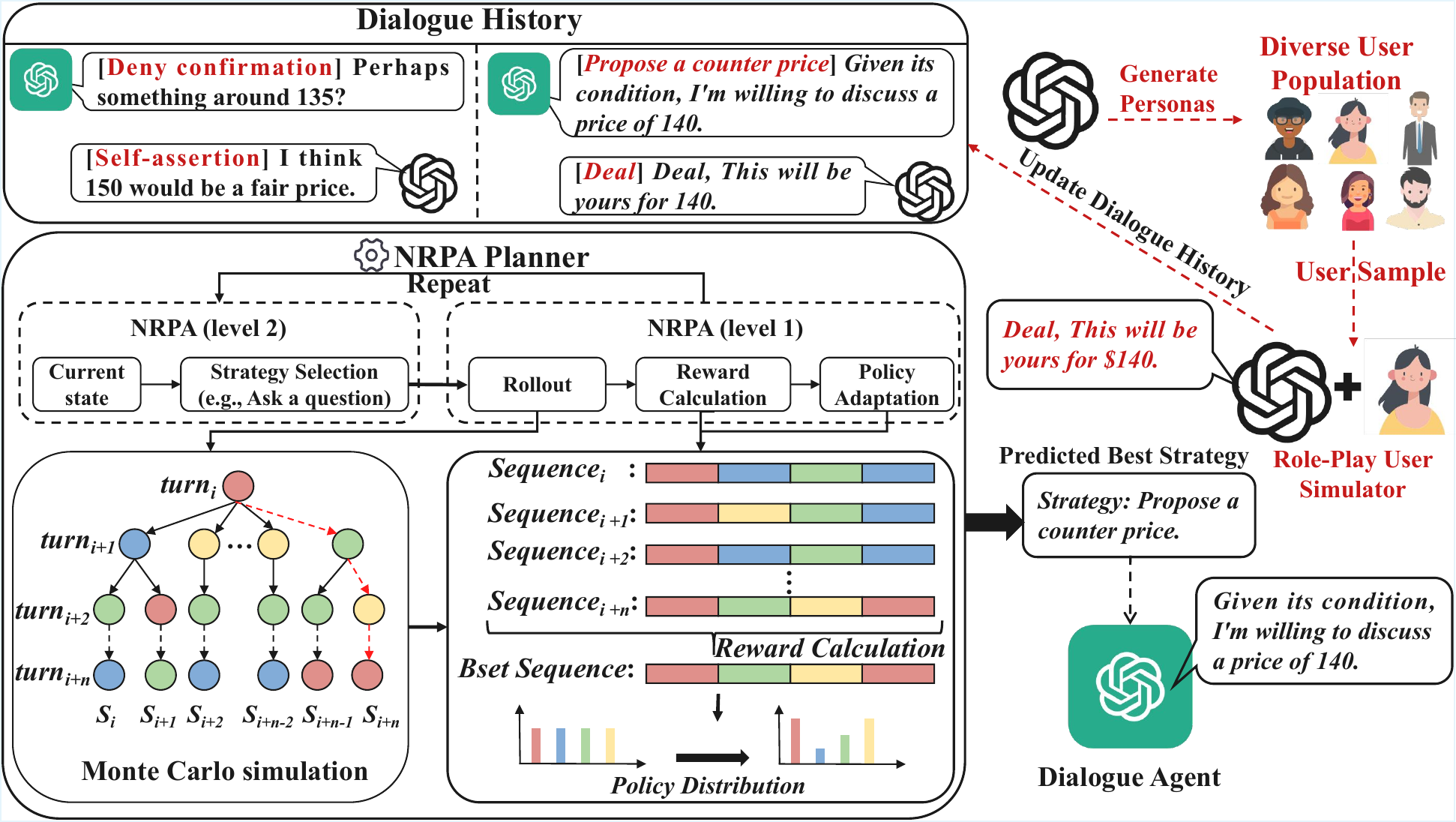}
    \caption{The overview of the UP-NRPA framework. This framework integrates a user portrait-driven simulator with a Nested Rollout Policy Adaptation planner. Through multi-level Monte Carlo simulation and reward-based policy adaptation, the agent dynamically optimizes dialogue strategies by simulating interactions with diverse user personas.}
    \label{fig:framework}
\end{figure*}

\section{Introduction}

With the development of Large Language Models (LLMs), goal-oriented dialogue systems have made substantial progress~\cite{LLM1,LLM2,dialog}. These systems excel in scenarios where goals align with user interests, such as restaurant reservations, emotional support, or multi-step task guidance~\cite{dilog2,li2024review,deng2024towards}, demonstrating robust collaborative dialogue capabilities. However, their performance significantly degrades when dialogue objectives conflict with user interests, such as in negotiation or persuasion scenarios~\cite{dilog2}. Therefore, the system must balance goal achievement and user sentiment in conversations to achieve optimal interaction outcomes.

With the advancement of dialogue systems, numerous novel solutions have emerged \cite{deng2024towards}. Approaches based on prompt engineering optimize decision-making processes by directly guiding model planning through well-crafted instructions and contextual prompts, or by integrating external policy planners to collaborate with LLMs, thereby constructing efficient dialogue agents \cite{ANE}. However, while these approaches show progress, they have limitations. Offline reinforcement learning methods perform well for single users but suffer from poor generalization capabilities, with performance heavily dependent on high-quality dialogue data, leading to costly training. Online search methods like Monte Carlo Tree Search (MCTS) can generate natural responses but fail to achieve objectives in goal-oriented dialogue tasks \cite{gdpzero}. Existing approaches fall short in modeling user personas. In real-world dialogue scenarios, each user possesses unique personality traits, yet current methods show poor performance in integrating these personas. Offline reinforcement learning struggles to train policy planners capable of generalizing across diverse user populations, resulting in dialogue agents exhibiting rigid behavioral strategies when encountering different users. In complex multi-user scenarios like persuasion \cite{P4G}, negotiation \cite{CB}, or emotional support \cite{ESconv,EXTES}, existing dialogue agents lack dynamic adaptability to adjust strategies based on user feedback. This limitation constrains their effectiveness in applications requiring deep interaction, empathy building, and trust establishment. Furthermore, existing approaches fail to maintain dialogue coherence and goal-oriented focus in non-collaborative tasks. They cannot capture behavioral shifts across different users and adapt conversational strategies accordingly. Collectively, these issues constrain the performance of LLM-based dialogue systems.

To tackle the challenges in dialogue planning posed by diverse user portraits, several innovative solutions have been proposed. Among them, the Tailored Strategic Planning (TRIP) \cite{trip} method achieves effective adaptation to personalized dialogue scenarios by deeply integrating user characteristics into the strategy planning process. Its innovation lies in integrating a user perception strategy planning module with a group-based training paradigm, systematically enhancing the agent's customized strategy planning capabilities and overcoming the limitations of traditional methods under single user simulators. Building upon this foundation, User-Tailored Dialogue Policy Planning (UDP)~\cite{udp} further expands the TRIP framework by employing advanced diffusion models to dynamically infer and model user profiles. This approach introduces a Brownian Bridge-inspired mechanism, enabling precise prediction of users' response patterns and behavioral tendencies. UDP not only captures real-time shifts in user characteristics, but also dynamically adjusts policy planning during conversations, achieving significant performance improvements over TRIP.

Although TRIP and UDP methods have made progress in dialogue systems based on user portraits, enabling strategy formulation for different user characteristics, both approaches still exhibit limitations in dialogue task performance metrics. They demonstrate relatively poor performance in terms of dialogue success rate and dialogue turns, requiring more rounds of interaction to achieve dialogue targets. However, the Nested Rollout Policy Adaptation for Goal-oriented Dialogue (NRPA-GD) \cite{NRPA-GD} method employs a policy adaptation mechanism to achieve efficient dialogue planning within a single user simulator environment, significantly improving dialogue success rates. Therefore, we propose the User Portrait-based Nested Rollout Policy Adaptation (UP-NRPA) method. This approach integrates user characteristics into nested rollout planning, dynamically adjusting dialogue strategies through online search and user feedback. This enables UP-NRPA to adaptively customize strategies across diverse user scenarios, effectively addressing the need for personalized dialogue planning across different user populations. Our contributions are summarized as follows:

\begin{itemize}
   \item Current dialogue strategy planning requires offline reinforcement learning, which cannot dynamically adjust strategies in real-time for unseen user personas. In contrast, UP-NRPA dynamically plans strategies through real-time feedback without requiring model training.
     \item UP-NRPA combines user profiling with online strategy optimization, enabling the system to continuously enhance interaction strategies based on distinct user personas, thereby improving dialogue success rates.
     \item In collaborative and non-collaborative dialogue tasks, UP-NRPA achieved a 100\% success rate across multiple tasks. Based on the Qwen2.5 14B model, this method improved the success rate by 56.41\% compared to existing state-of-the-art approaches, validating the effectiveness of UP-NRPA.
\end{itemize}

\section{ Related Work}

Existing prompt engineering methods, such as Ask-an-Expert (AnE), integrate active prompting, self-reflection, and self-play to enhance LLMs' planning capabilities by learning from context and history through predefined instruction prompts~\cite{ANE,chen2023controllable,ICL-AIF}. However, these prompting techniques typically prioritize user satisfaction. User simulations often employ fixed neutral roles, struggling to reflect individual characteristics~\cite{deng2023rethinking}. Plug-and-Play Dialogue Policy Planner (PPDPP)~\cite{ppdpp} and Dual-Process Dialogue Planning (DPDP)~\cite{dpdp} achieve stronger policy capabilities through offline reinforcement learning. Additionally, TRIP~\cite{trip} and UDP~\cite{udp} methods incorporate user profiles to identify user types and adopt customized dialogue strategies. However, these approaches remain reliant on substantial data and offline training, limiting their ability to adapt strategies for new users during dialogue planning. Latent Dialogue Policy Planning (LDPP)~\cite{ldpp} is based on data-driven autonomous policy discovery. In dialogue scenarios such as negotiation, persuasion, and emotional support, the interaction process contains rich contextual information. The system must autonomously adjust dialogue policies to achieve predetermined interaction goals. LDPP implements the entire process from policy mining in dialogue records to policy planning learning. Based on an offline hierarchical reinforcement learning algorithm in latent space, it constructs efficient policy planning capabilities. Goal-oriented Dialogue Planning with Zero training (GDP-Zero) utilizes LLMs to simultaneously process prior strategies, value functions, and user/system roles, enabling MCTS planning for unknown scenarios~\cite{gdpzero}. NRPA, a variant of Nested Monte Carlo Search, is applied to goal-oriented dialogue systems via multi-level policy adaptation~\cite{wang2025planning}. NRPA-GD addresses the computational overhead of offline reinforcement learning by introducing an online NRPA search algorithm, thereby improving dialogue success rates. DialogXpert generates a small set of high-quality action candidates for each dialogue turn using frozen LLM models. It then leverages a compact Q-network based on fixed BERT embeddings trained via temporal difference learning to select optimal actions within a reduced feature space. By tracking user sentiment, DialogXpert advances tasks while making customized decisions to establish genuine empathetic connections~\cite{dialogxpert}. The proposed UP-NRPA model combines user role modeling with nested rollout, iteratively optimizing optimal action sequences based on user feedback to enable dialogue agent to select optimal strategies.

\begin{table*}[!t]
\centering
\begin{tabular}{@{}llccccc@{}}
\toprule
\multirow{2}{*}{Method} & \multirow{2}{*}{Backbone} & \multicolumn{3}{c}{CraigslistBargain} & \multicolumn{2}{c}{ESConv} \\
\cmidrule(lr){3-5} \cmidrule(l){6-7}
 & & AT $\downarrow$ & SR $\uparrow$ & SL $\uparrow$ & AT $\downarrow$ & SR $\uparrow$ \\
\midrule
DialoGPT \cite{dialogpt} & GPT-2 & 6.73 & 0.3245 & 0.2012 & 5.31 & 0.7538 \\
\midrule
Standard \cite{ANE} & - & 6.47 & 0.3830 & 0.1588 & 5.10 & 0.7692 \\
AnE & - & 5.91 & 0.4521 & 0.2608 & 4.76 & 0.8000 \\
Proactive \cite{Proactive} & - & 5.80 & 0.5638 & 0.2489 & 5.08 & 0.7538 \\
\hspace{1em}+ MI-Prompt \cite{ppdpp} & - & 5.74 & 0.5691 & 0.2680 & 4.78 & 0.7846 \\
ProCoT \cite{Proactive} & - & 6.22 & 0.5319 & 0.2486 & 4.75 & 0.7923 \\
\hspace{1em}+ MI-Prompt \cite{ppdpp} & - & 6.12 & 0.5532 & 0.3059 & 4.83 & 0.7769 \\
ICL-AIF \cite{ICL-AIF} & - & 6.53 & 0.3617 & 0.1881 & 4.69 & 0.8079 \\
\midrule
\midrule
PPDPP \cite{ppdpp}& Vicuna 13B & 5.62 & 0.6117 & 0.3376 & 4.56 & 0.8462 \\
\hspace{1em}-w/o SFT & & 5.71 & 0.6223 & 0.3354 & 4.68 & 0.8384 \\
\hspace{1em}-w/o RL & & 5.57 & 0.6649 & 0.2280 & 5.24 & 0.7308 \\
\midrule
\midrule
DPDP (System 1) \cite{dpdp} & GPT-3.5-Turbo & 5.03 & 0.7447 & 0.4108 & 3.61 & 0.9000 \\
\hspace{1em}-System 1 w/o PT & & -- & -- & -- & 4.22 & 0.8769 \\
\hspace{1em}-System 1 w/o SPT & & -- & -- & -- & 3.97 & 0.8692 \\
\hspace{1em}-System 2 & & 2.78 & 0.9734 & 0.2728 & 2.13 & 0.9923 \\
\hspace{1em}-System 1 \& 2 & & -- & -- & -- & \textbf{2.13} & 0.9923 \\

\midrule
\midrule
% TRIP \cite{trip}&GPT-4o mini & 6.34 & 0.6888 & 0.409 & -- & -- \\
% \midrule
% \midrule
UDP \cite{udp} & GPT-4o mini & -- & -- & -- & 7.59 & 0.8320 \\
\hspace{1em}-w/o PT & & -- & -- & -- & 7.48 & 0.7720 \\
\hspace{1em}-w/o RL & & -- & -- & -- & 8.64 & 0.5310 \\
\midrule
\midrule
NRPA-GD level 1 \cite{NRPA-GD}  & GPT-4o mini & 3.89 & 0.9894 & 0.6326 & 5.28 & 0.9461 \\            
NRPA-GD level 2 & & 2.72& \textbf{1.0000} & 0.6422 & 4.17 & \textbf{1.0000} \\
\midrule
\midrule
DialogXpert \cite{dialogxpert}& Vicuna 13B & 2.93 & 0.9415 & 0.3811 & 2.7 & 0.9651 \\
\hspace{1em}-w/o RL & & 5.13 & 0.7561 & 0.3473 & 4.13 & 0.8749 \\
\midrule
DialogXpert & Qwen 1.8B & 2.78 & 0.9274 & 0.3791 & 2.49 & 0.9805 \\
\hspace{1em}-w/o RL & & 4.69 & 0.7754 & 0.3012 & 4.04 & 0.8921 \\
\midrule
DialogXpert & Qwen2.5 14B & \textbf{2.32} & 0.9746 & 0.4389 & 2.31 & 0.9876 \\
\hspace{1em}-w/o RL & & 3.64 & 0.8754 & 0.2952 & 3.53 & 0.9401 \\
\hspace{1em}-w/o LLM-Prior & & 3.31 & 0.9165 & 0.3598 & 3.89 & 0.9243 \\
\hspace{1em}-w/o Emotion & & 2.75 & 0.9136 & 0.3156 & 3.08 & 0.9611 \\
\midrule
\midrule
UP-NRPA level 1(ours)  & GPT-4o mini & 5.22 & 0.9096 & 0.7882 & 3.81 & \textbf{1.0000} \\            
UP-NRPA level 2(ours) & & 4.74& 0.9787 & \textbf{0.9069} & 3.25 & \textbf{1.0000} \\
\midrule
UP-NRPA level 1(ours) & Qwen2.5 14B & 3.14 & \textbf{1.0000} & 0.5835 & 2.92 & \textbf{1.0000} \\            
UP-NRPA level 2(ours) & & 2.88 & \textbf{1.0000} & 0.6865 & 2.76 & \textbf{1.0000} \\
\bottomrule
\end{tabular}
\caption{Comparison of dialogue planning methods on the CraigslistBargain, ESConv benchmarks.}
\label{tab:1}
\end{table*}

\begin{algorithm}[!t]
\caption{UP-NRPA}
\label{alg:nrpa-complete}
\begin{algorithmic}[1]
\Require LLM $M_\theta$
\Require Initial policy $\pi$
\Require Number of iterations $N$
\Require Learning rate $\alpha$
\Require Action space $\mathcal{A}$
\Require Initial state $s$
\Require User Portrait $U$
\Function{UP-NRPA}{$level$, $\pi$, $s$}
    \If{$level = 0$}
        \State \Return \Call{Playout}{$s$, $\pi$, $U$}
    \Else
        \State $bestScore \gets -\infty$
        \State $bestSequence \gets \emptyset$
        
        \For{$iteration= 1$ to $N$}
            \State $(score, sequence) \gets$ \Call{NRPA}{$level-1$, $\pi$, $s$}
            
            \If{$score > bestScore$}
                \State $bestScore \gets score$
                \State $bestSequence \gets sequence$
            \EndIf
            
            \State $\pi \gets$ \Call{Adapt}{$\pi$, $bestSequence$, $\alpha$, $s$}
        \EndFor
        
        \State \Return $(bestScore, bestSequence)$
    \EndIf
\EndFunction
\Function{Adapt}{$\pi$, $sequence$, $\alpha$, $s$}
    \State $\pi' \gets \pi$
    \State $currentState \gets s$
    
    \For{each action $a$ in $sequence$}
        \State $z \gets \sum_{a' \in \mathcal{A}} e^{\pi'(a')}$
        
        \For{each action $a' \in \mathcal{A}$}
            \State $\pi'(a') \gets \pi'(a') - \alpha \cdot \frac{1}{z} e^{\pi'(a')}$
        \EndFor
        
        \State $\pi'(a) \gets \pi'(a) + \alpha$
        \State $currentState \gets play(currentState, a)$
    \EndFor
    
    \State \Return $\pi'$
\EndFunction
\end{algorithmic}
\end{algorithm}

\section{User-Specific Planning Evaluation}
To address the challenge that existing dialogue strategy planning methods struggle to dynamically adapt to diverse user characteristics, this paper proposes the UP-NRPA online framework to explore its planning adaptation capabilities. Two task categories are examined: collaborative and non-collaborative tasks. Collaborative tasks include the cooperative dialogue tasks ESConv and ExTES for emotional support scenarios \cite{ESconv,EXTES}, while non-collaborative tasks encompass the non-cooperative dialogue task P4G for persuasion scenarios \cite{P4G} and the CB non-cooperative dialogue task for negotiation scenarios \cite{CB}. First, user profiles are customized using Big Five personality traits and decision-making styles. Then, GPT-5 generates user descriptions with fine-grained characteristics based on these profiles. Finally, comparative experiments validate the task performance of UP-NRPA.

\subsection{User Persona Designing}
Building upon existing research, we combine persona portraits with user simulation and select non-cooperative behaviors from a set of resisting strategies when dealing with non-cooperative tasks. The generation and integration process of persona portraits follows the research framework of TRIP~\cite{trip}, similarly designing two role types, each equipped with coherent setting descriptions generated by large language models. These descriptions encompass two key dimensions: Big Five personality traits \cite{goldberg1992development} and decision-making styles \cite{scott1995decision}. Simultaneously, we employ the resisting strategy proposed by \cite{dutt2021resper} to guide simulator behavior patterns. Hybrid active role-playing prompts designed for each agent integrate specific character settings with dialouge context information. For each evaluation task, we constructed 300 diverse user simulators to ensure comprehensive and systematic test coverage. For details on the resisting strategy, see Appendix E.

\section{Methodology}
\subsection{Problem Definition}
Given existing research, the dialogue planning process can be formalized as a Markov Decision Process (MDP) \cite{trip,dialogxpert}, represented as the tuple $(\mathcal{S}, \mathcal{A}, \mathcal{R}, \mathcal{T})$, where $S$ denotes the dialogue state space, $\mathcal{A}$ denotes the dialogue action space, $\mathcal{R}$ is the reward function, and $\mathcal{T}$ is the state transition function. At each dialogue time step $t$, the dialogue state $s_t \in \mathcal{S}$ encompasses the complete dialogue context and historical record. The agent selects an action $a_t \in \mathcal{A}$ based on the current state, triggering a state transition $s_{t+1} = \mathcal{T}(s_t, a_t)$ and receiving an immediate reward $\mathcal{R}_t$. The core objective of the dialogue agent is to learn an optimal policy. The reward function $\mathcal{R}$ is designed based on the NRPA-GD reward mechanism, calculating rewards according to the dialogue termination state ($1$ or $0$), dialogue turn number, and corresponding penalty terms \cite{NRPA-GD}. The output space of this planner is a predefined set of policies based on existing research, where each policy is accompanied by pre-designed natural language instruction descriptions. 

\begin{table*}[ht]
\centering
% \resizebox{\columnwidth}{!}{
    \begin{tabular}{@{}l l c c c c@{}}
    \toprule
    Method & Backbone & \multicolumn{2}{c}{P4G} & \multicolumn{2}{c}{ExTES} \\
    \cmidrule(lr){3-4} \cmidrule(l){5-6}
     & & AT $\downarrow$ & SR $\uparrow$ & AT $\downarrow$ & SR $\uparrow$ \\
    \midrule
    Standard & - & 8.32 & 0.468 & -- & -- \\
    ProCoT \cite{Proactive} & - & 7.98 & 0.543 & -- & -- \\
    ICL-AIF \cite{ICL-AIF}& - & 8.06 & 0.465 & 7.65 & 0.555 \\
    GDP-Zero \cite{gdpzero}& - & 9.12 & 0.328 & -- & -- \\
    TRIP \cite{trip}& GPT3.5 & 8.20 & 0.495 & -- & -- \\
    \midrule
    PPDPP \cite{ppdpp}& Vicuna 13B & 8.185 & 0.463 & 8.163 & 0.558 \\
    \midrule
    UDP \cite{udp}& GPT-4o mini & 7.705 & 0.598 & -- & -- \\
    \hspace{1em}- w/o PT & & 8.017 & 0.513 & -- & -- \\
    \hspace{1em}- w/o RL & & 8.000 & 0.533 & -- & -- \\
    \midrule
    LDPP \cite{ldpp}& Qwen1-1.8B & 5.57 & 0.795 & 4.132 & 0.903 \\
    \hspace{1em}- w/o 2nd Stage & & 6.14 & 0.760 & 4.483 & 0.865 \\
    \hspace{1em}- w/o 3rd Stage & & 6.84 & 0.570 & 7.038 & 0.623 \\
    \midrule
    DialogXpert \cite{dialogxpert}& Vicuna 13B & 5.07 & 0.8132 & 2.97 & 0.9534 \\
    \midrule
    DialogXpert & Qwen1-1.8B & 3.97 & 0.8793 & 2.73 & 0.9651 \\
    \midrule
    DialogXpert & Qwen2.5 14B & 3.34 & 0.9129 & \textbf{2.57} & 0.9782 \\
    \midrule
    \midrule
    UP-NRPA level 1(ours) & GPT-4o mini & 4.91 & 0.9747 & 4.71 & 0.9900 \\
    UP-NRPA level 2(ours) & & 4.36 & \textbf{1.0000} & 4.22 & \textbf{1.0000} \\
    \midrule
    UP-NRPA level 1(ours) & Qwen2.5 14B & 3.40 & 0.9184 & 3.29 & \textbf{1.0000} \\
    UP-NRPA level 2(ours) & & \textbf{3.12} & \textbf{0.9849} & 2.69 & \textbf{1.0000} \\
    \bottomrule
    \end{tabular}
% }
\caption{Comparison of dialogue planning methods on the P4G, ExTES benchmarks.}
\label{tab:2}
\end{table*}

\subsection{UP-NRPA}
The proposed UP-NRPA framework integrates user modeling with online search algorithms to achieve real-time policy optimization in goal-oriented dialogue. As shown in Figure \ref{fig:framework}, the system first employs the diverse user population for sampling, constructing structured user profiles. These profiles then drive the Role-Play User Simulator. Within the NRPA Planner, the system employs a nested search mechanism for online planning. At level 2, the agent preliminarily selects a policy based on the current dialogue state. Subsequently, it proceeds to level 1, where monte carlo simulation is used to conduct multiple turns of complete dialogue simulation. During this process, the user simulator provides feedback on the agent's actions based on the predefined portrait. Based on reward calculations, continuously update the policy distribution. Without relying on offline reinforcement learning training, optimal policy output is achieved in complex scenarios. In algorithm \ref{alg:nrpa-complete}, the UP-NRPA process begins at the nested hierarchy level, recursively searching for improved action sequences that maximize dialogue rewards. It leverages LLMs to generate system responses and user portrait driven user replies, appending dialogue pairs to the state until termination conditions are met, thereby simulating a complete dialogue trajectory. Through reward computation, if a candidate result outperforms the current optimal outcome, the optimal score and sequence are updated, and policy $\pi$ adjusts toward the new optimal sequence. For more implementation details, see Appendix A.

\section{Experimental Setup}
\subsection{Evaluation Tasks}
We evaluated the performance of the proposed method on collaborative and non-collaborative tasks. Specifically, on the CraigslistBargain (CB) \cite{CB}, evaluations were conducted using 3,290 training samples, 188 validation samples, and 188 test samples containing bargaining dialogues between buyers and sellers. For ESConv \cite{ESconv}, which focuses on emotional support, we used 1,040 training samples, 130 validation samples, and 130 test samples. For P4G \cite{P4G}, centered on donation persuasion, we employed 817 training samples, with 100 samples each for validation and testing. Additionally, the expanded version of ESConv, ExTES, was utilized \cite{EXTES}. This dataset contains richer data, comprising 10,717 training samples, 200 validation samples, and 200 test samples. Since the UP-NRPA method does not require offline reinforcement learning, it was evaluated directly on the test set. For these four dialogue tasks, Appendix D provides a more detailed introduction.

\subsection{Evaluation Metrics}
Following previous studies  \cite{dpdp}, we selected Average Turns (AT) and Success Rate (SR). AT measures the efficiency of goal completion by calculating the average dialogue turns required to reach the target. SR measures the effectiveness of goal completion by statistically determining the percentage of goals successfully achieved within a predetermined maximum number of turns. The Sale-to-List Ratio (SL) is used to evaluate buyers' transaction outcomes. A higher SL value indicates greater buyer benefit from the transaction, if the transaction fails, SL is recorded as 0. Its calculation formula is defined as: \textit{ SL\% = (deal price - seller target price)/(buyer target price - seller target price)}. Additionally, we introduce the Soft Success Rate (SSR) evaluation method proposed by LDPP \cite{ldpp} to further assess the effectiveness of UP-NRPA. SSR serves as a complementary enhancement to SR, which binarily maps the final turn rewards of a dialogue to determine only task success or failure. In contrast, SSR averages all final turn rewards directly. Taking P4G as an example, persuasion success is rated as: \textit{refused} $\rightarrow$ -1.0, \textit{neutral} $\rightarrow$ -0.5, \textit{positive inclination} $\rightarrow$ 0.1, and \textit{agreed to donate} $\rightarrow$ 1.0. Detailed information about ESConv task is provided in Appendix C.

\subsection{Baselines}
Dialogue models based on fine-tuning technology are represented by DialoGPT \cite{dialogpt}, a pre-trained dialogue generation model whose core function is to automatically generate natural, coherent, and information-rich responses given a dialogue context. Prompt engineering approaches, such as Standard Prompt \cite{dpdp}, drive LLMs to generate responses through foundational prompts; Proactive \cite{Proactive} and ProCoT \cite{Proactive} introduce explicit goal planning chains within prompts, Ask-an-Expert~\cite{ANE} simulates expert standard reasoning strategies through predefined prompts, while ICL-AIF \cite{ICL-AIF} generates text feedback for context learning without parameter updates via model self-play. GDPZero \cite{gdpzero} employs MCTS to find optimal solutions. Offline reinforcement learning-based approaches PPDPP\cite{ppdpp} and DPDP \cite{dpdp} combine offline reinforcement learning with real-time MCTS search optimization. TRIP \cite{trip} incorporates user portraits and Theory-of-Mind (ToM) to simulate more realistic scenarios. UDP \cite{udp} builds upon TRIP by using diffusion models to construct user portraits and predicting user feedback via a Brownian-bridge mechanism. LDPP \cite{ldpp} employs a Variational Autoencoder (VAE) to extract latent strategies from real dialogues, then offline-trains a hierarchical strategy planner within this latent space. DialogXpert \cite{dialogxpert} employs a Q-network trained on BERT embeddings for rapid optimal decision-making.

\begin{figure}[ht]
    \centering
    \begin{subfigure}{0.495\linewidth}
        \includegraphics[width=\linewidth]{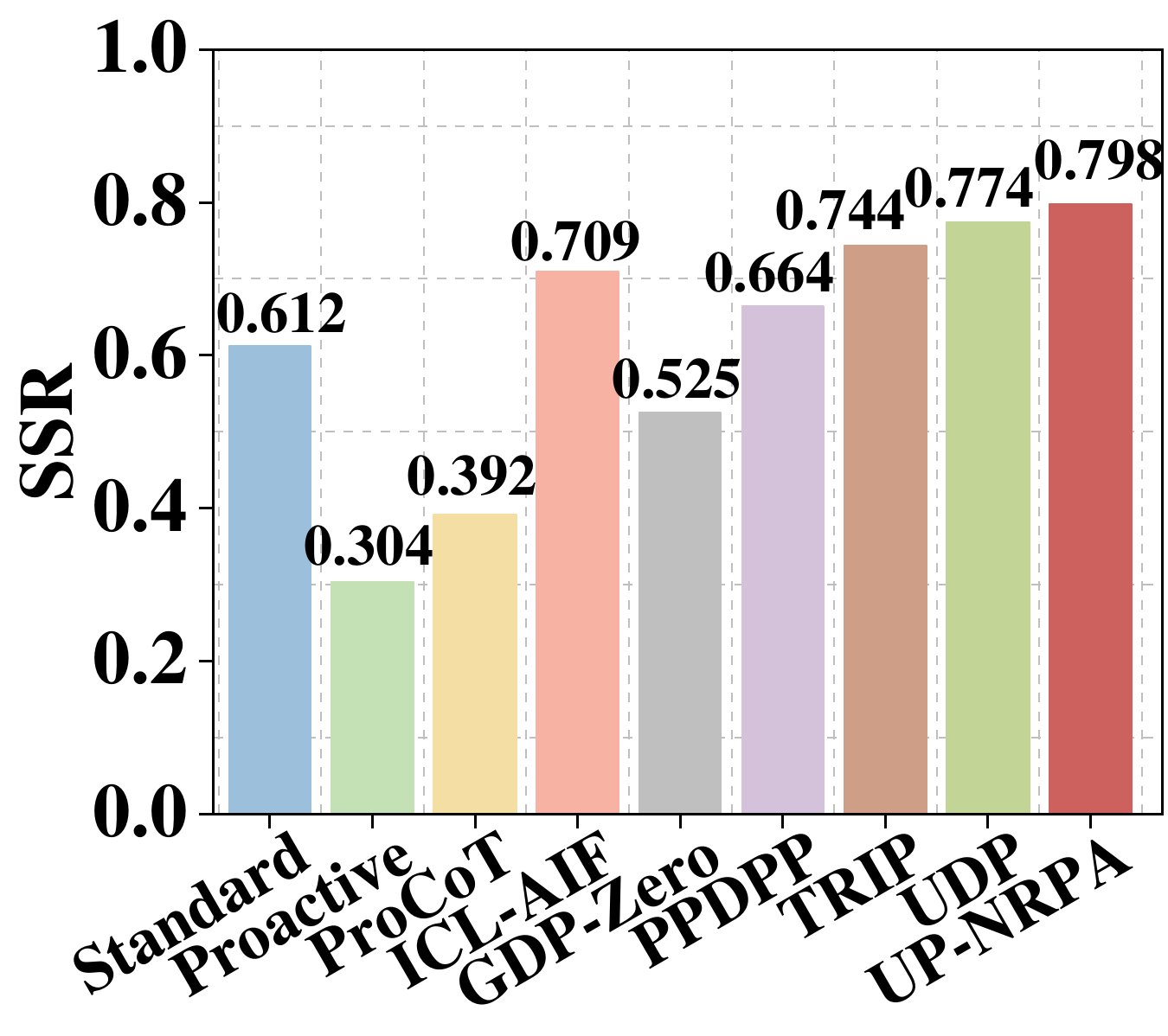}
        \caption{ESConv}
        \label{fig:esconv_ssr}
    \end{subfigure}
    \hfill
    \begin{subfigure}{0.495\linewidth}
        \includegraphics[width=\linewidth]{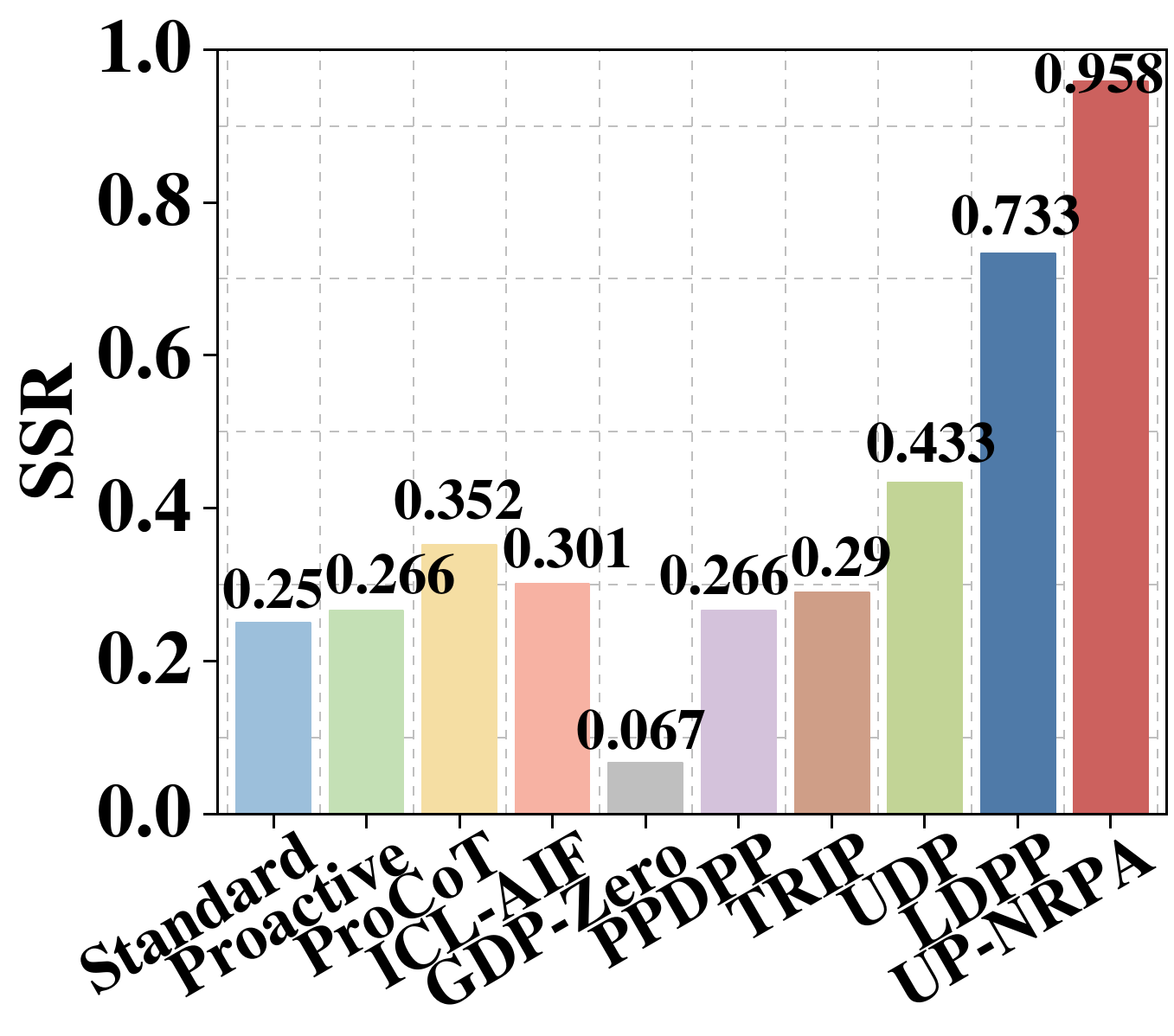}
        \caption{P4G}
        \label{fig:P4G_ssr}
    \end{subfigure}
    \caption{Comparison of different methods' performance on SSR using ESConv and P4G.}
    \label{fig:ssr}
\end{figure}

\section{ Experiments and Analysis}
\subsection{Main Results}
Table \ref{tab:1} presents performance comparisons of various methods across multiple challenging dialogue planning benchmarks. We evaluated the proposed UP-NRPA planner on both collaborative dialogue planning benchmarks (ESConv, ExTES) and non-collaborative dialogue planning benchmarks (CB, P4G). This summary encompasses diverse baseline approaches, including various planners and recent policy-based language model methods. The UP-NRPA planner was tested using Qwen 2.5 14B and GPT-4o-mini backbone models. The experimental setup references the user portrait construction approach of the TRIP method and employs a user simulator with resistance policies on non-collaborative tasks to construct complex, realistic interaction scenarios. Although UP-NRPA performed slightly worse on GPT-4o-mini than on Qwen2.5-14B, It achieved the highest results on SL, improving by approximately \textbf{41.22\%} over NRPA-GD at level 2. In contrast, DialogXpert was evaluated on Qwen2.5-14B in a relatively simpler environment. Our approach achieves exceptionally high SR across both tasks. For CB, while AT slightly outperforms DialogXpert in more complex settings, our method demonstrates superior performance on the critical negotiation metrics SR and SL. It achieves SR of \textbf{1.0000} at both levels and elevates SL from 0.4389 to \textbf{0.6856} at level 2. In the ESConv task, it still maintained an SR of \textbf{1.0000}, despite slightly more dialogue turns.

Table \ref{tab:2} presents experimental results for the P4G and ExTES tasks, demonstrating performance differences among various methods. On the P4G task, the TRIP method using the backbone GPT-3.5 achieves AT = 8.20 and SR = 0.495. Meanwhile, the UDP method using the backbone GPT-4o-mini achieves AT = 7.705 and SR = 0.598 on P4G. As the current state-of-the-art baseline, DialogXpert achieves AT = 3.34 and SR = 0.9129 on the P4G task using Qwen2.5 14B, and attains AT = 2.57 and SR = 0.9782 on the ExTES task. Our proposed UP-NRPA planner, based on Qwen2.5 14B, achieves SR = 0.9184 and AT = 3.40 for Level 1 on the P4G task, and SR = \textbf{1.0000} with AT=3.29 on the ExTES task. Level 2 further optimizes AT to 3.12 and improves SR to 0.9849 on the P4G task; it maintains SR = \textbf{1.0000} while achieving AT = 2.69 on the ExTES task. Compared to methods like TRIP and UDP that utilize user portraits, our UP-NRPA method demonstrates better performance across all key metrics. Compared to DialogXpert, which does not use user portraits, our method achieves higher SR values, indicating that user portraits contribute to improved dialogue generation performance.

We compared the SSR performance of different methods on ESConv and P4G, as shown in Figure \ref{fig:ssr}. On the ESConv, UP-NRPA achieved an SSR = \textbf{0.798}, outperforming the previously well-performing UDP (0.774) and TRIP (0.744). On the non-collaborative task P4G, UP-NRPA demonstrated even more pronounced advantages, achieving an SSR =\textbf{ 0.958} significantly outperforming all comparison baselines, including LDPP (0.733). This indicates that in non-cooperative tasks, the method can stably steer conversations toward desired reward objectives, substantially enhancing the overall performance and task completion quality of dialogue systems.

\begin{figure}[ht]
    \centering
    \begin{subfigure}{0.495\linewidth}
        \includegraphics[width=\linewidth]{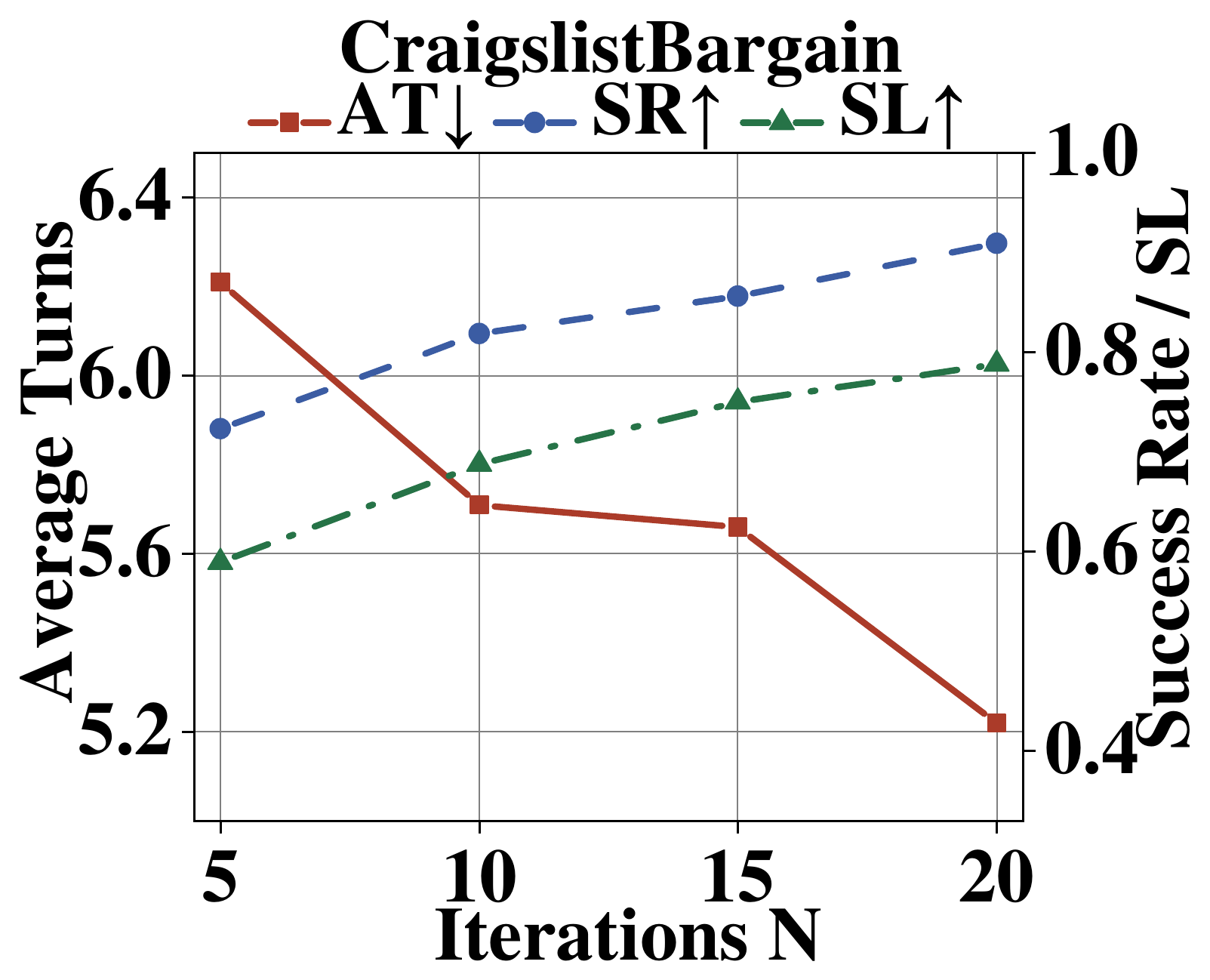}
        \label{fig:CB_Ablation}
    \end{subfigure}
    \hfill
    \begin{subfigure}{0.495\linewidth}
        \includegraphics[width=\linewidth]{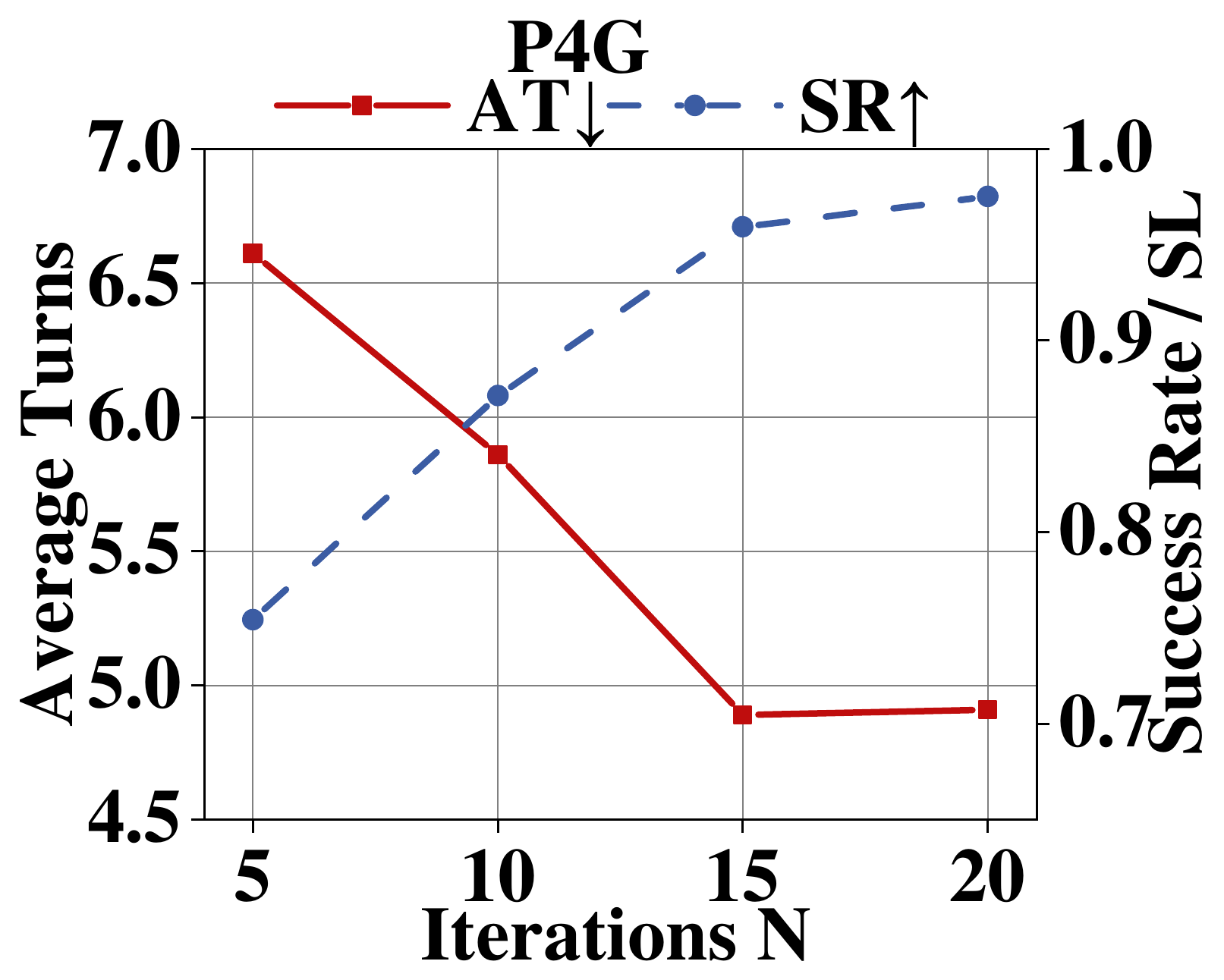}
        \label{fig:P4G_Ablation}
    \end{subfigure}
    \caption{Performance of non-collaborative tasks at different iteration N in the level 1 of UP-NRPA.}
    \label{fig:cb and p4g}
\end{figure}

\subsection{Ablation Study}
This section analyzes the impact of iteration N on UP-NRPA through ablation studies to validate the algorithm's performance in both collaborative and non-collaborative tasks.

\subsubsection{Analysis of Iteration N}
The ablation results for UP-NRPA level 1 on two non-collaborative tasks are shown in Figure \ref{fig:cb and p4g}. The overall performance of the UP-NRPA planner exhibits a significant positive correlation with its iteration N. As N increases, AT shows a substantial downward trend across both tasks, while SR and SL metrics steadily improve. At N = 5, the SR values on CB and P4G were 0.7234 and 0.7542, respectively. As N increased to 20, the SR values reached 0.9096 and 0.9750, respectively. This validates that increasing the number of iterations for UP-NRPA effectively enhances the model's exploration capability within the search space, thereby achieving higher-quality negotiation outcomes within fewer dialogue turns.

For collaborative tasks, as shown in Table \ref{tab:collaborative task}, we conducted ablation experiments on both the ESConv and ExTES tasks. Similar to the non-collaborative tasks in Figure \ref{fig:cb and p4g}, the number of iterations N also positively impacts performance. For the collaborative task UP-NRPA, an SR of 1.0000 was achieved on the ESConv task at N = 5. As N increases, accuracy drops to 3.76 at N = 10, yielding the optimal test result. As an extension of the ESConv dataset, ExTES exhibits similar behavior to the ESConv task, achieving SR $\geq$ 0.98 as early as level 1. We observe that N = 10 maintains high SR values while avoiding excessive simulation time consumption. This demonstrates that UP-NRPA significantly enhances the success rate of collaborative dialogues.

\begin{table}[ht]
\centering
\begin{tabular}{@{}l c c c c@{}}
\toprule
& \multicolumn{2}{c}{ESConv} & \multicolumn{2}{c}{ExTES} \\
\cmidrule(lr){2-3} \cmidrule(l){4-5}
Method  & AT $\downarrow$ & SR $\uparrow$ & AT $\downarrow$ & SR $\uparrow$ \\
\midrule
UP-NRPA(N= 5)  & 3.95 & \textbf{1.0000} &4.89 & \textbf{0.9950} \\
UP-NRPA(N=10)  & \textbf{3.76} & \textbf{1.0000} & 4.69 & 0.9800 \\
UP-NRPA(N=15)& \textbf{3.76} & \textbf{1.0000} & 4.74 & \textbf{0.9950} \\
UP-NRPA(N=20)  & 3.81 & \textbf{1.0000} & \textbf{4.71} & 0.9900 \\
\bottomrule
\end{tabular}
\caption{Performance of collaborative tasks at different iteration N in the level 1 of UP-NRPA.}
\label{tab:collaborative task}
\end{table}

% \begin{table}[h!]
% \centering
% \resizebox{\columnwidth}{!}{
% \begin{tabular}{@{}l c c c c c c c c@{}}
% \toprule
% UP-NRPA& \multicolumn{2}{c}{Ide.} & \multicolumn{2}{c}{Com.} & \multicolumn{2}{c}{Sugg.} & \multicolumn{2}{c}{Ove.}\\
% \cmidrule(lr){2-3} \cmidrule(lr){4-5} \cmidrule(lr){6-7} \cmidrule(lr){8-9}
% vs. & Win& Lose  & Win & Lose  & Win & Lose  & Win & Lose \\
% \midrule
% NRPA-GD  & 1 & 2 & 4.89 & \textbf{0.9950} & 3 & 4 & 5 & 6 \\
% \bottomrule
% \end{tabular}
% }
% \label{tab:ESCONV AND EXTES}
% \caption{Human evaluation results on ExTES and P4G.}

% \end{table}

\begin{figure}[ht]
    \centering
    \begin{subfigure}{0.495\linewidth}
        \includegraphics[width=\linewidth]{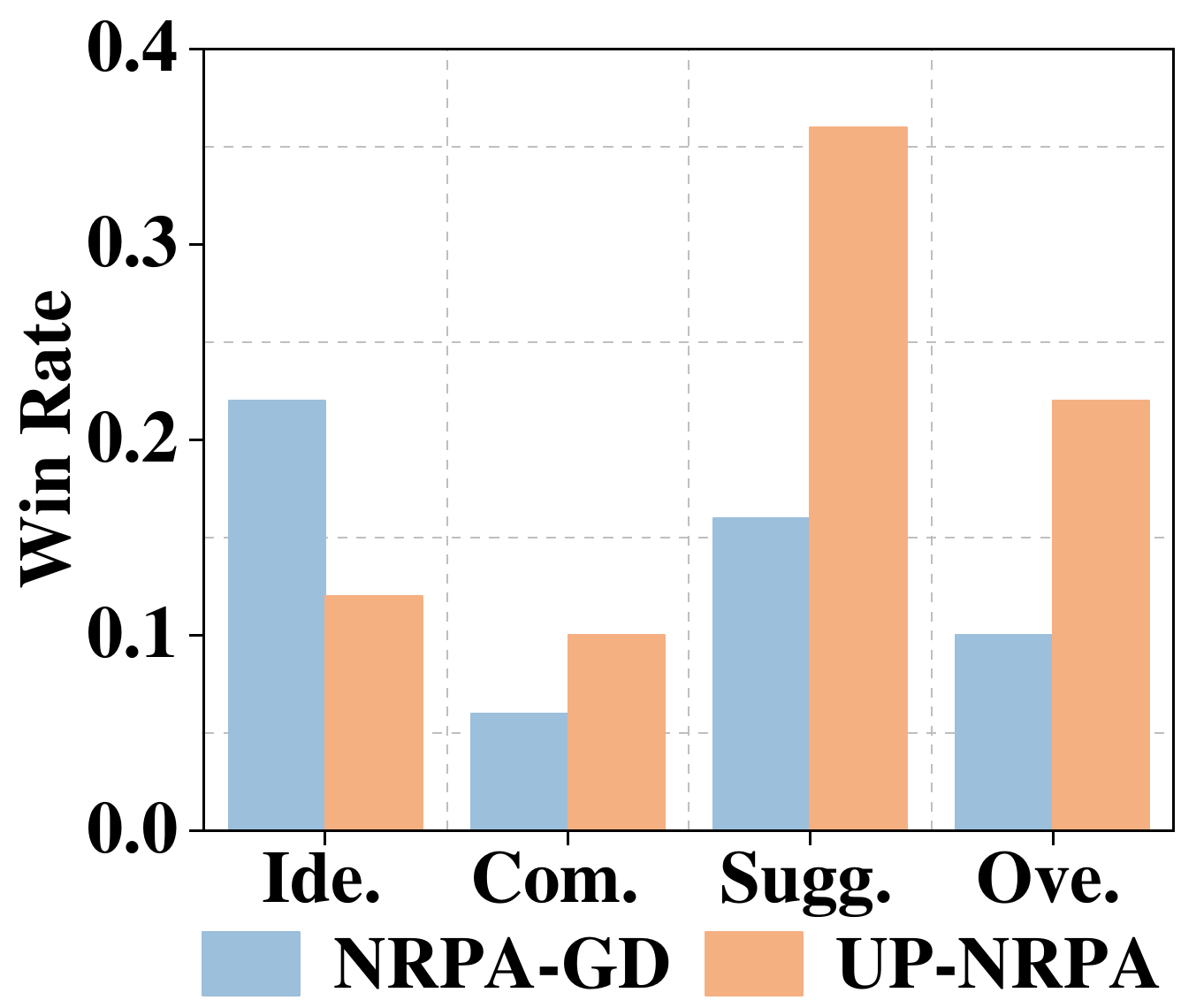}
        \caption{ESConv}
        \label{fig:esconv_human}
    \end{subfigure}
    \hfill
    \begin{subfigure}{0.495\linewidth}
        \includegraphics[width=\linewidth]{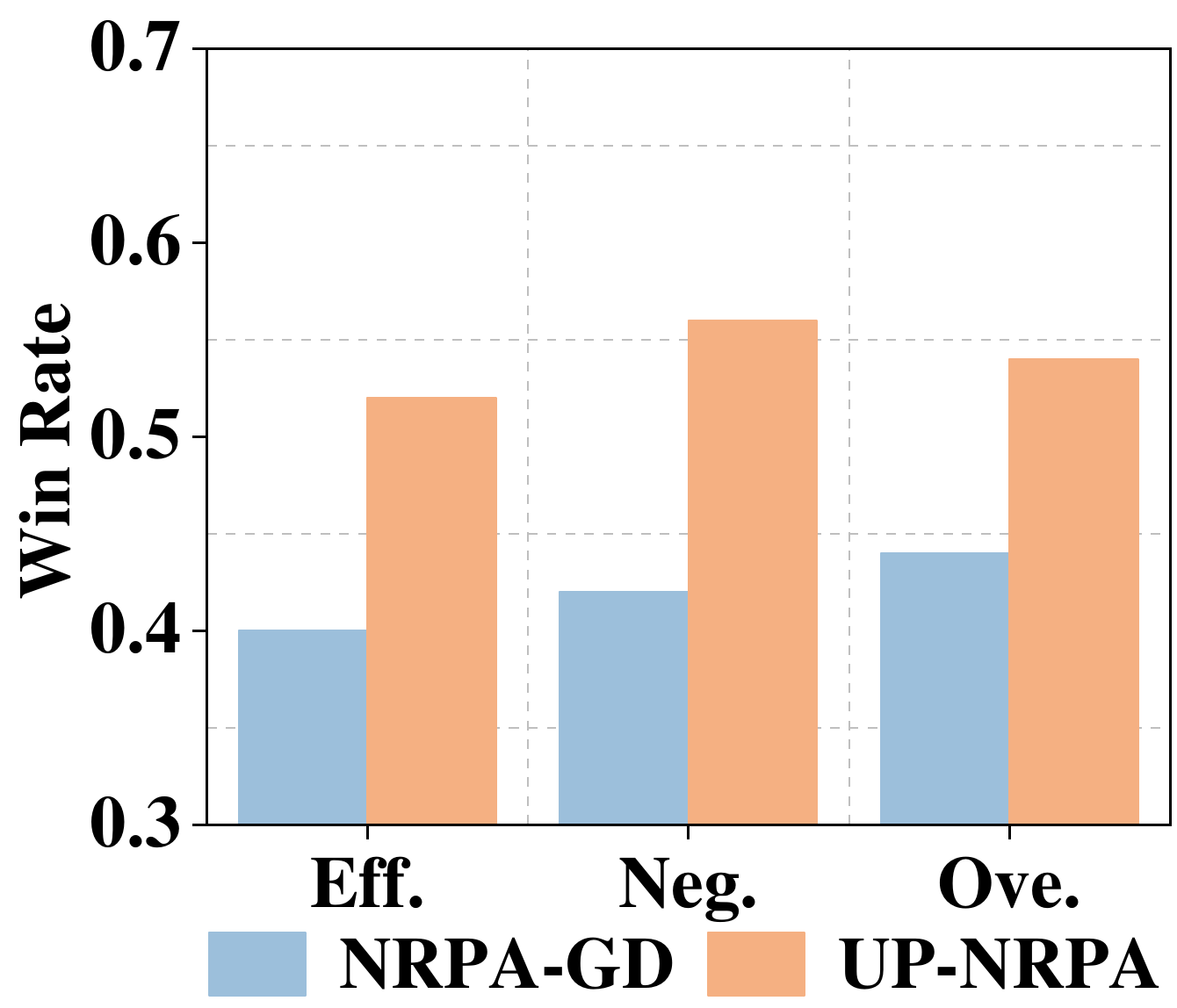}
        \caption{CraigslistBargain}
        \label{fig:CraigslistBargain_human}
    \end{subfigure}
    \caption{Human evaluation results on ESConv and CraigslistBargain.}
    \label{fig:human}
\end{figure}

\subsection{Human Evaluation}
We conducted human evaluations of responses generated by UP-NRPA and NRPA-GD. Based on prior research \cite{ppdpp,NRPA-GD}, 50 test samples are randomly selected from ESConv and CB. Three annotators independently assessed responses generated on ESConv across four dimensions: Suggestion (Sug., comparing the quality of suggestions), Identification (Ide., comparing the proactivity in addressing emotional issues), Comforting (Com., comparing the quality of comfort), and Overall (Ove.). For responses generated on CB, the dimensions are Effectiveness (Eff., comparing the effectiveness in achieving negotiation outcomes), Negotiation (Neg., comparing the strength of negotiation skills and tactics), and Overall (Ove., comparing the overall negotiation capabilities). As shown in Figure \ref{fig:human}, on the ESConv task,	UP-NRPA demonstrated superior ability to provide high-quality suggestions, outperforming NRPA-GD in Overall. On the CB task, UP-NRPA surpassed NRPA-GD across all metrics. Aligned with the results in Table \ref{tab:1}, UP-NRPA particularly excelled in SL, indicating its outperformance over NRPA-GD in negotiation. This further validates the effectiveness of UP-NRPA in both collaborative and non-collaborative dialogue tasks. We also compared UP-NRPA's level 1 (N = 20) and level 2 (N = 5) on the P4G and ExTES tasks to evaluate performance differences across levels. Detailed human evaluation and results are provided in Appendix B.
\section{Conclusion}
This paper introduces the UP-NRPA framework to address the limitations of existing dialogue strategy planning systems in adapting to diverse user tasks and achieving low success rates. By integrating user modeling with strategy adaptation, the framework enables real-time policy adjustment without relying on offline reinforcement learning. During conversations, UP-NRPA dynamically optimizes dialogue strategies in real time by responding to user feedback, thereby enhancing the performance of dialogue agents in both collaborative and non-collaborative tasks while demonstrating particular strengths in negotiation scenarios. This framework offers a viable solution for developing dialogue systems adaptable to diverse user types. For future work, we plan to further optimize computational efficiency in complex dialogue scenarios and extend the framework for application in multimodal dialogue environments.

\section*{Acknowledgments}
The authors acknowledge the financial support from National Natural Science Foundation of China, No. 62236002, and  Hefei Key Science and Technology Special Projects under Grant 2024SZD006.

\bibliographystyle{named}
\bibliography{ijcai26}

\clearpage
\setcounter{section}{0}
\renewcommand{\thesection}{\Alph{section}}
\section{Implementation Details}
\subsubsection{Experimental Details}
In our experimental setup, we employed prompts from TRIP and NRPA-GD, implementing the framework based on NRPA-GD. To ensure comparability and consistency across experiments, we adhered to established research standards for dataset selection: ESConv, P4G, ExTES, and CraigslistBargain datasets strictly adhered to Dialogxpert's test set divisions. For model configuration, both the dialogue system and user model employed Qwen2.5-14B and GPT-4o-mini as backbone models. Critical hyperparameters such as temperature settings were maintained identical to NRPA-GD's original configuration to ensure fairness and reproducibility of experimental results.

\subsubsection{Nested Rollout Policy Adaptation}
The NRPA algorithm modifies the sampling mechanism in the MCTS simulation phase by integrating online policy learning within the recursive framework of Nested Monte Carlo Search (NMCS). Traditional methods employ fixed probabilities or predefined rules for rollouts, whereas NRPA maps the action space to weighted parameters and utilizes a Boltzmann distribution to generate action probabilities. At each nested level, the algorithm uses high-reward sequences obtained through search as supervisory signals. It then adjusts the parameter distribution via gradient ascent, biasing sampling toward historically optimal paths. This mechanism eliminates the need for explicit node storage, instead guiding search space convergence through weight evolution. It achieves a shift in simulation policy from static distribution to dynamic feedback-driven adaptation. Specifically, let the state of the first \(t\) step be \(s_t\), and let the set of legitimate actions be denoted as \(\mathcal{A}(s_t)\). We parameterize the strategy as a vector 
: $\pi \in \mathbf{R}^{|\mathcal{A}|}$, 
where the component \(\pi(a)\) corresponds directly to the weight of action \(a\). Given the optimal sequence of actions \((a_1,a_2,\dots,a_T)\) for a high score rollout, the following updates are performed for each step \(t\):

Calculate the softmax normalization factor.
\begin{equation} 
z = \sum_{a'\in\mathcal{A}} e^{\pi(a')}
\end{equation}

Calculate the probability of each action.
\begin{equation} 
P(a)=\frac{e^{\pi(a)}}{z}
\end{equation}

Update the weights for all actions \(a'\in\mathcal{A}\), and add an extra \(\alpha\) to the optimal action \(a\). 
\begin{equation} 
\begin{cases}
\pi(a')\gets\pi(a')-\alpha \cdot \frac{1}{z}e^{\pi(a')}, & \forall a' \in \mathcal{A} \\
\pi(a)\gets\pi(a)+\alpha
\end{cases}
\end{equation}

The net increase in weight of the optimal action \(a\) is \(\alpha - \alpha \cdot \frac{1}{z}e^{\pi(a)} = \alpha(1-P(a))\), and the net decrease in weight of the remaining actions is \(\alpha \cdot \frac{1}{z}e^{\pi(a')} = \alpha \cdot P(a')\), which transforms the original random simulation that was performed blindly into an adaptive sampling that continuously concentrates on the optimal direction.

\section{Human Evaluation Details}
To evaluate the quality of responses generated by this model, we organized a controlled human evaluation in accordance with LDPP protocols, inviting three expert annotators with backgrounds in natural language processing and computer science to participate. Each annotator reviewed 50 dialogue scenarios. Through a majority voting mechanism among the three annotators, preference results for each dimension were aggregated item by item. This evaluation process ensures our assessment of the quality of both emotional support dialogues and negotiation dialogues is grounded in professional expertise. As shown in Figure \ref{fig:human}, we compared different levels of UP-NRPA. Level 2 demonstrated superiority in two tasks, but its effect was not significant in the emotional support task.

\begin{figure}[!ht]
    \centering
    \begin{subfigure}{0.495\linewidth}
        \includegraphics[width=\linewidth]{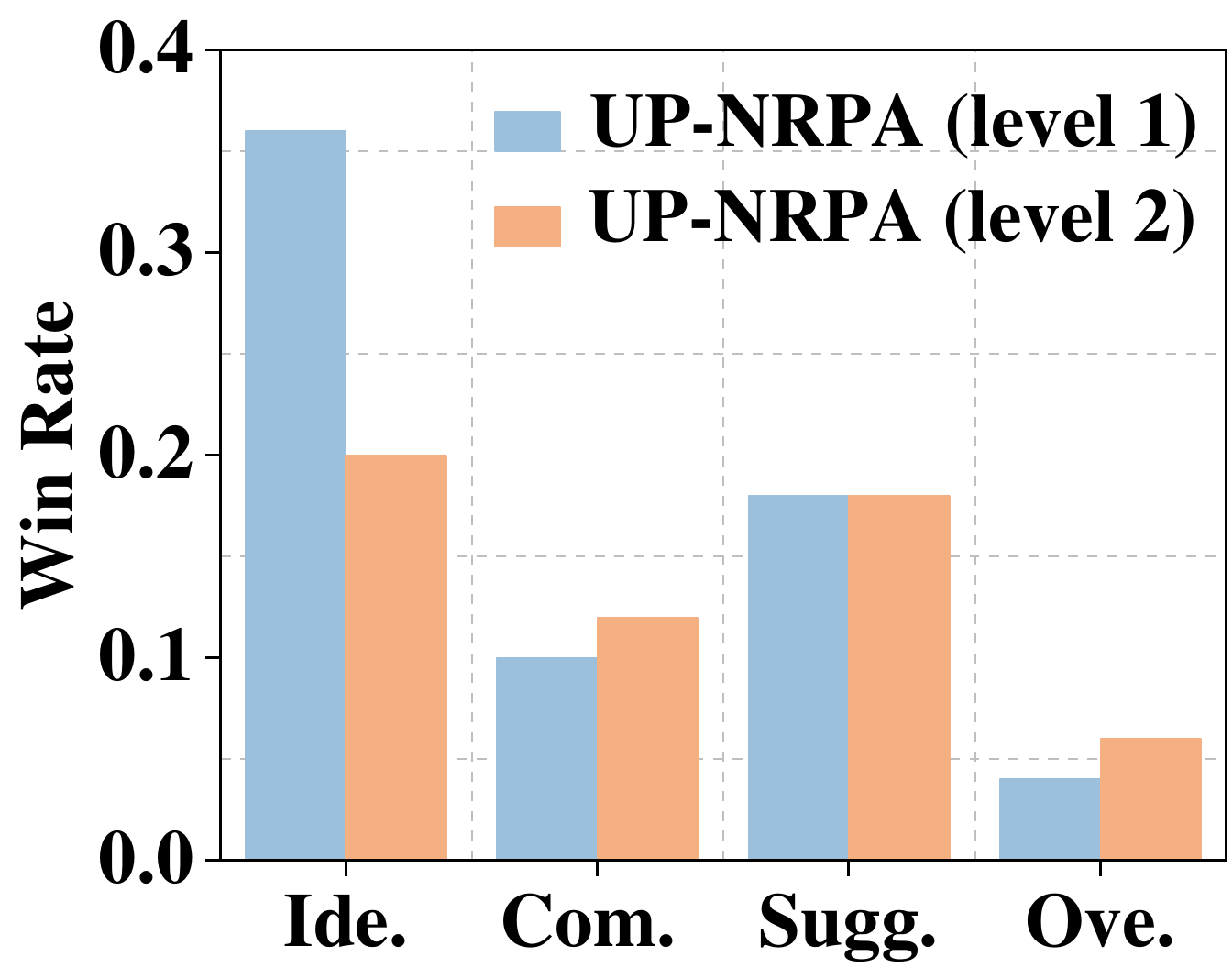}
        \caption{ExTES}
        \label{fig:ExTES}
    \end{subfigure}
    \hfill
    \begin{subfigure}{0.495\linewidth}
        \includegraphics[width=\linewidth]{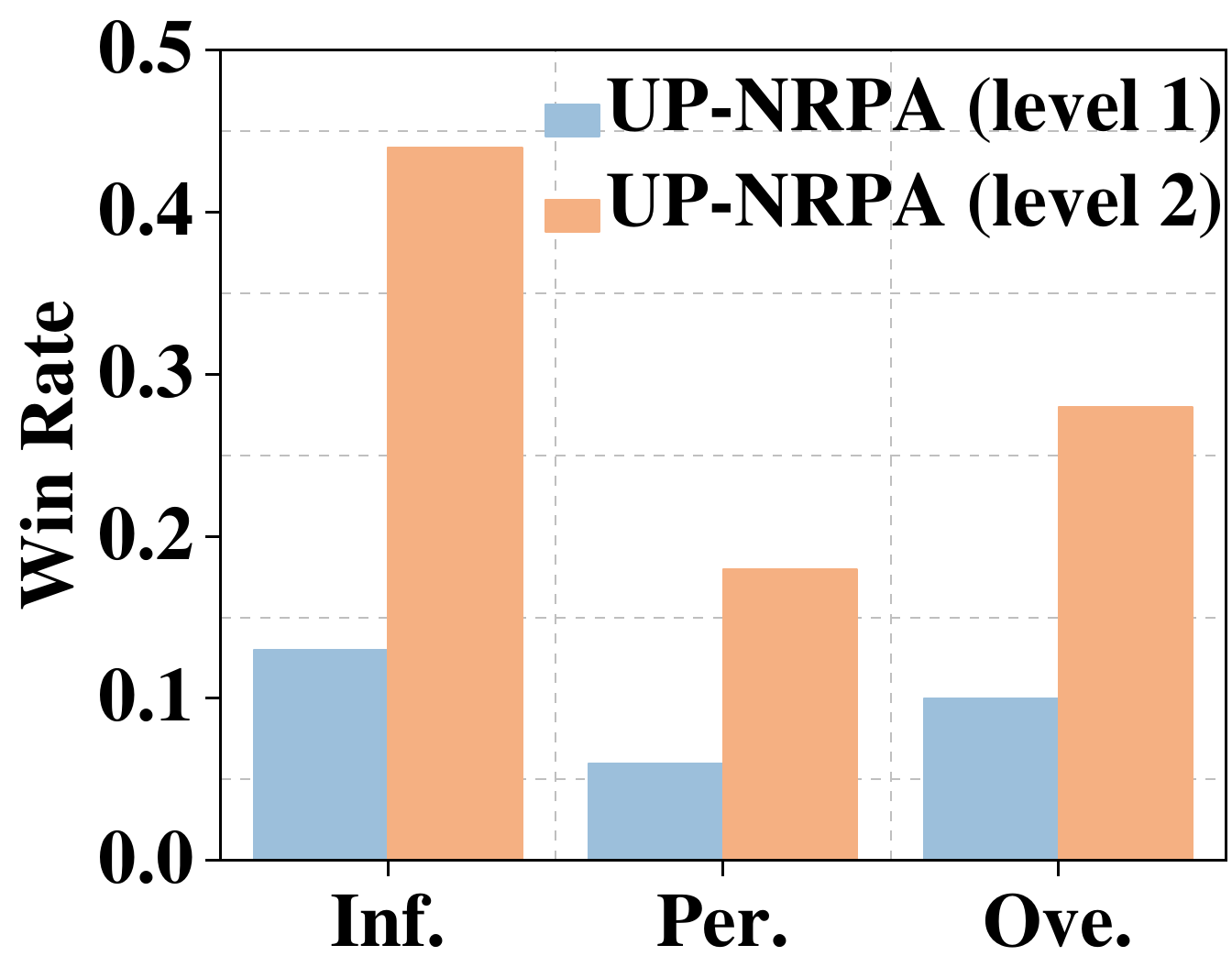}
        \caption{P4G}
        \label{fig:P4G}
    \end{subfigure}
    \caption{Human evaluation results on ExTES and P4G.}
    \label{fig:human}
\end{figure}
As for ExTES, we measure four main metrics of the generated dialogues as follows:
\begin{itemize}
    \item \textbf{Identification:} Which assistant is more helpful in exploring and identifying the problem?
    \item \textbf{Comforting:} Which assistant is more skilled at comforting you?
    \item \textbf{Suggestion:} Which assistant provides more helpful suggestions for solving the problem?
    \item \textbf{Overall:} Which assistant can better solve the patient's problem?
\end{itemize}

As for P4G, we measure three main metrics of the responses as follows:
\begin{itemize}
    \item \textbf{Informative:} Which assistant's introduction to the charity was more engaging?
    \item \textbf{Persuasive:} Which assistant takes the more persuasive approaches?
    \item \textbf{Overall:} Which assistant has stronger persuasive capabilities?
\end{itemize}

\begin{table*}[h]
\centering
\begin{tabularx}{\textwidth}{@{}l X X@{}}
\toprule
\textbf{Resisting Strategy} & \textbf{Persuasion (P4G)} & \textbf{CraigslistBargain (CB)} \\
\midrule
Source Derogation & Attacks/doubts the organisation's credibility. & Attacks the other party or questions the item. \\
\midrule
Counter Argument & Argues that the responsibility of donation is not on them or refutes a previous statement. & Provides a non-personal argument/factual response to refute a previous claim or to justify a new claim. \\
\midrule
Personal Choice & Attempts to saves face by asserting their personal preference such as their choice of charity and their choice of donation. & Provides a personal reason for disagreeing with the current situation or chooses to agree with the situation provided some specific condition is met. \\
\midrule
Information Inquiry & Ask for factual information about the organisation for clarification or as an attempt to stall. & Requests for clarification or asks additional information about the item or situation. \\
\midrule
Self Pity & Provides a self-centred reason for not being able/willing to donate at the moment. & Provides a reason (meant to elicit sympathy) for disagreeing with the current terms. \\
\midrule
Hesitance & Attempts to stall the conversation by either stating they would donate later or is currently unsure about donating. & Stalls for time and is hesitant to commit; specifically, they seek to further the conversation and provide a chance for the other party to make a better offer. \\
\midrule
Self-assertion & Explicitly refuses to donate without even providing a factual/personal reason. & Asserts a new claim or refutes a previous claim with an air of finality/confidence. \\
\midrule
Others & Do not explicitly foil the persuasion attempts. & Do not explicitly foil the negotiation attempts. \\
\bottomrule
\end{tabularx}
\caption{The resisting strategies for P4G and CB tasks.}
\label{tab:resisting_strategies}
\end{table*}

\section{Detailed Evaluation of Soft Success Rate}
According to the SSR proposed by LDPP, SR calculates success rates by mapping final-round rewards to binary values of 0 or 1, while SSR directly averages all final-round rewards. Therefore, we regard SSR as a soft success rate metric. Each dataset employs a task-specific reward mapping scheme.
\begin{itemize}
    \item \textbf{P4G}: Persuasion success is rated as: 
    $\text{refused} \rightarrow -1.0$, $\text{neutral} \rightarrow -0.5$, 
    $\text{positive inclination} \rightarrow 0.1$, and $\text{agreed to donate} \rightarrow 1.0$.
    \item \textbf{ESConv}: Emotion trajectories are scored as follows: 
    $\text{worse} \rightarrow -1.0$, $\text{same} \rightarrow -0.5$, $\text{better} \rightarrow 0.5$, and $\text{solved} \rightarrow 1.0$.    
\end{itemize}

These mappings enable consistent supervision across diverse tasks while adapting to domain-specific success criteria.

\section{Dataset Details}

\begin{itemize}
    \item \textbf{ESConv}: Emotional support and therapy. The goal, as a therapist, is to help the patient resolve their emotional issues.
    
    \item \textbf{CB}: Negotiating for price haggle. Roleplaying as the buyer in the conversation, the goal is to buy a given product as close as possible to the buyer's target price in order to maximize profit.
    
    \item \textbf{ExTES}: Emotional support and therapy. Similar to ESConv but more diverse and larger in sample size. The goal, as a therapist, is to help the patient resolve their emotional issues.
    
    \item \textbf{P4G}: Persuasion for donation. The goal, as a role player, is to persuade a persuadee to donate to a charity called ``Save the Children''.
\end{itemize}
\section{Resisting Strategy}
We employ the resisting strategies proposed in the TRIP study to simulate users' non-cooperative behavior. Table \ref{tab:resisting_strategies} provides detailed descriptions of these resisting strategies.

\section{Comprehensive Prompting}
By integrating character descriptions with resistance strategies, we constructed a comprehensive prompt framework for our user simulator. Specifically, our prompts comprise two components: task context and dialogue history. Within the task context, we guide the large language model to simulate designated characters through role-playing instruction sets and resistance strategies. Tables \ref{tab:user_simulator_prompt}, \ref{tab:charity_simulator_prompt} and \ref{tab:emotional_simulator_prompt} respectively present complete user simulator prompt templates for the two task categories.

\begin{table*}[h]
\centering

\begin{tabularx}{\textwidth}{@{}X@{}}
\toprule
\textbf{The prompt for user persona generation} \\
\midrule
You need to select one attribute from each of the following persona types. \\
\\
Persona types \\
Big-Five Personality: ["openness", "conscientiousness", "extraversion", "agreeableness", "neuroticism"] \\
Decision-Making Styles: ["directive", "analytical", "conceptual", "behavioral"] \\
\\
Please generate a list of N fictional user profiles. \\
\bottomrule
\end{tabularx}
\caption{The prompt of user persona generation.}
\label{tab:persona_prompt}
\end{table*}

\begin{table*}[h]
\centering
\begin{tabularx}{\textwidth}{@{}X@{}}
\toprule
\textbf{The prompt for user persona rephrase} \\
\midrule
You need to incorporate the following persona attributes and generate a cohesive persona description. You need to ensure the description is easy to understand. \\
\\
Big-Five Personality: \\
Decision-Making Style: \\
\\
\midrule
An Example: You are a 28-year-old female software developer. Your personality is characterized by openness to experience, which means you are curious, imaginative, and willing to try new things. In your occupation, you excel at analyzing problems and finding logical solutions. Your decision-making style is analytical, meaning you carefully consider all available information before making a choice. \\
\bottomrule
\end{tabularx}
\caption{The prompt of user persona rephrase.}
\label{tab:persona_rephrase_prompt}
\end{table*}

\begin{table*}[h]
\centering
\begin{tabularx}{\textwidth}{@{}X@{}}
\toprule
\textbf{The user simulator prompt for the price bargain task} \\
\midrule
Now enter the role-playing mode. In the following conversation, you will play as a seller in a price bargaining game. Your persona: \textless Persona Description\textgreater \\
\\
You must follow the instructions below during chat. \\
1. Your utterances and bargain behavior need to strictly follow your persona. Varying your wording and avoid repeating yourself verbatim! \\
2. You can decide to change your target price flexibly based on your persona and the conversation. \\
\\
Your Response Strategy: \\
1. "Source Derogation": Attacks the other party or questions the item. \\
2. "Counter Argument": Provides a non-personal argument/factual response to refute a previous claim or to justify a new claim. \\
3. "Personal Choice": Provides a personal reason for disagreeing with the current situation or chooses to agree with the situation provided some specific condition is met. \\
4. "Information Inquiry": Requests for clarification or asks additional information about the item or situation. \\
5. "Self Pity": Provides a reason (meant to elicit sympathy) for disagreeing with the current terms. \\
6. "Hesitance": Stalls for time and is hesitant to commit; specifically, they seek to further the conversation and provide a chance for the other party to make a better offer \\
7. "Self-assertion": Asserts a new claim or refutes a previous claim with an air of finality/ confidence. \\
8. "Others": Do not explicitly foil the negotiation attempts. \\
\\
You are the seller who is trying to sell the \%s with the initial price of \%s. Product description: \%s. Please reply with only one short and succinct sentence. \\
\bottomrule
\end{tabularx}
\caption{The comprehensive prompt of user simulators in the price negotiation task.}
\label{tab:user_simulator_prompt}
\end{table*}

\begin{table*}[h]
\centering
\begin{tabularx}{\textwidth}{@{}X@{}}
\toprule
\textbf{The user simulator prompt for the charity persuasion task} \\
\midrule
Now enter the role-playing mode. In the following conversation, you will play as a Persuadee in a persuasion game. Your persona: \textless Persona Description\textgreater \\
\\
You must follow the instructions below during chat. \\
1. Your utterances need to strictly follow your persona. Varying your wording and avoid repeating yourself verbatim! \\
2. Pretend you have little knowledge about the Save the Children charity. You have little willingness for donation at the beginning of conversation. \\
3. Your willingness for donation depends on your persona and how influenced you are by the Persuader. \\
4. You need to determine whether to donate money or not. If you think your are convinced by the Persuader, you should donate money for the charity. \\
\\
Your Response Strategy: \\
1. "Donate": show your willingness to donate. \\
2. "Source Derogation": attacks or doubts the organisation's credibility. \\
3. "Counter Argument": argues that the responsibility is not on them or refutes a previous statement. \\
4. "Personal Choice": Attempts to saves face by asserting their personal preference such as their choice of charity and their choice of donation. \\
5. "Information Inquiry": Ask for factual information about the organisation for clarification or as an attempt to stall. \\
6. "Self Pity": Provides a self-centred reason for not being willing to donate at the moment. \\
7. "Hesitance": Attempts to stall the conversation by either stating they would donate later or is currently unsure about donating. \\
8. "Self-assertion": Explicitly refuses to donate without even providing a personal reason. \\
9. "Others": Do not explicitly foil the persuasion attempts. \\
\\
You are the Persuadee who is being persuaded by a Persuader. Please reply with only one short and succinct sentence. \\
\bottomrule
\end{tabularx}
\caption{The comprehensive user simulator prompt for the charity persuasion task.}
\label{tab:charity_simulator_prompt}
\end{table*}

\begin{table*}[h]
\centering
\begin{tabularx}{\textwidth}{@{}X@{}}
\toprule
\textbf{The user simulator prompt for the emotional support task} \\
\midrule
Now enter the role-playing mode. In the following conversation, you will play as a Patient in a counselling conversation with a therapist. Your persona: \textless Persona Description\textgreater \\
\\
You must follow the instructions below during chat. \\
1. Your utterances need to strictly follow your persona. Varying your wording and avoid repeating yourself verbatim!\\
2. You are seeking help from the therapist for your emotional issue.\\
3. Your emotional state and response style depends on your persona.\\
\\
Your Response Strategy: \\
1. "Feel worse": You feel worse after the therapist's response.\\
2. "Feel the same": No change in your emotional state.\\
3. "Feel better": You feel somewhat better but issue not fully resolved.\\
4. "Solved": Your emotional issue has been fully resolved and you feel much better.\\
\\
You are the patient who is looking for the help from the therapist. Please reply with only one short and succinct sentence.. \\
\bottomrule
\end{tabularx}
\caption{The comprehensive user simulator prompt for the emotional support task.}
\label{tab:emotional_simulator_prompt}
\end{table*}

\end{document}